\documentclass{article}

\usepackage{microtype}
\usepackage{graphicx}
\usepackage{subfigure}
\usepackage{multirow}
\usepackage{amsmath}
\usepackage{amsthm}
\usepackage{amssymb}
\usepackage{tikz}
\usepackage{lipsum}
\usepackage{systeme}
\usepackage{cases}
\usepackage{enumitem}
\usepackage{mathtools}
\usepackage{booktabs} 
\usepackage[colorinlistoftodos,prependcaption,textsize=tiny]{todonotes}
\usepackage{hyperref}

\newtheorem{assumption}{Assumption}
\newcommand{\Cov}{\mathrm{Cov}}

\usepackage[accepted]{icml2019}

\icmltitlerunning{Structural Redundancy Reduction in Channel Pruning - A Theoretic Study}

\begin{document}
\twocolumn[
\icmltitle{Investigating Channel Pruning through Structural Redundancy Reduction - A Statistical Study}


\icmlsetsymbol{equal}{*}
\begin{icmlauthorlist}
\icmlauthor{Chengcheng Li*}{utk}
\icmlauthor{Zi Wang*}{utk}
\icmlauthor{Dali Wang}{utk,ornl}
\icmlauthor{Xiangyang Wang}{sysu}
\icmlauthor{Hairong Qi}{utk}
\end{icmlauthorlist}

\icmlaffiliation{utk}{University of Tennessee, Knoxville, TN, USA.}
\icmlaffiliation{ornl}{Oak Ridge National Laboratory, Oak Ridge, TN, USA.}
\icmlaffiliation{sysu}{Sun Yat-Sen University, Guangzhou, China}

\icmlkeywords{Machine Learning, ICML}
\vskip 0.3in
]
\printAffiliationsAndNotice{\icmlEqualContribution}
\begin{abstract}
Most existing channel pruning methods formulate the pruning task from a perspective of inefficiency reduction which iteratively rank and remove the least important filters, or find the set of filters that minimizes some reconstruction errors after pruning. In this work, we investigate the channel pruning from a new perspective with statistical modeling. We hypothesize that the number of filters at a certain layer reflects the level of ``redundancy'' in that layer and thus formulate the pruning problem from the aspect of redundancy reduction. Based on both theoretic analysis and empirical studies, we make an important discovery: randomly pruning filters from layers of high redundancy outperforms pruning the least important filters across all layers based on the state-of-the-art ranking criterion. These results advance our understanding of pruning and further testify to the recent findings that the structure of the pruned model plays a key role in the network efficiency as compared to inherited weights.
\end{abstract}


\section{Introduction}
\label{sec:intro}
Deep convolutional neural networks have achieved significant success in a wide range of studies \cite{hu2014convolutional, he2016deep, silver2017mastering, christiansen2018silico}. However, the property of over-parameterization limits their employment on resource-limited platforms and applications such as robotics, portable devices, and drones \cite{sandler2018mobilenetv2, ma2018shufflenet}. Channel pruning, by removing a whole set of filters as well as their corresponding feature maps, has been developed  as an important approach to improve the efficiency of neural networks  without customized software or hardware \cite{sze2017efficient}. In this work, we study saliency-based channel-pruning, a significant branch of channel pruning. 

Existing saliency-based approaches iteratively rank the importance of filters with certain criteria and remove the lowest-ranked (least important) filters. Taylor expansion estimates the loss change of each filter's removal \cite{molchanov2016pruning}. There are also heuristic criteria, e.g. minimum magnitudes of weights and the mean values of the activation maps \cite{li2016pruning} and \cite{polyak2015channel}. Although these kinds of methods achieved considerable pruning ratio while maintaining the performance, as we will show in our studies, there are still rooms for further improvement.

In this paper, we formulate channel pruning as a process of redundancy reduction to the network structure. With a statistical model, we prove that when a certain layer has much higher redundancy than other layers, randomly pruning filters from that layer can even outperform pruning filters with the lowest ranks across all layers. This finding is also verified by empirical studies. We manually increase the number of filters in a certain layer of AlexNet and VGG-16 and find that randomly pruning filters from the created redundant layer performs much better than using the state-of-the-art criterion-based approach, i.e., the Taylor expansion approach. We further show that a naive redundancy reduction based approach, which removes filters from the layer with the most filters in standard AlexNet and VGG-16, can outperform the Taylor expansion approach.

Based on the theoretic analysis and empirical studies, we believe that network structure obtained through iterative redundancy reduction, rather than a procedure of unimportant weights removal, plays a more significant role in better sustaining a network performance. Our findings imply that exploring the redundancy lying in a neural network is a promising research direction for future pruning study.

\section{Pruning as Redundancy Reduction - A Theoretic Analysis}
\label{sec:math}

\label{sec:analysis}
We formulate the problem of channel pruning from redundancy reduction perspective, where redundancy refers to the number of filters at a certain convolutional layer. 

Without loss of generality, suppose we have a two-layer DNN with $m$ and $n$ filters, respectively, where $n \gg m$. Let $\{\xi_1,\xi_2,\cdots,\xi_m\}$ and $\{\eta_1,\eta_2,\cdots,\eta_n\}$ be random variables, representing the contributions of all the filters in these two layers, respectively, and $\xi_i$ ($i=1,\cdots,m$)  and $\eta_i$ ($i=1,\cdots,n$) are positive scalars, i.e., $P(\xi_i>0)=1,P(\eta_i>0)=1$.

{\bf Claim:} If a certain layer has much higher redundancy than others, randomly pruning filters from that layer outperforms pruning filters with the lowest ranks across all layers. 

Denote $\underline{\xi}=\min\{\xi_1,...,\xi_m\}$, $\underline{\eta}=\min\{\eta_1,...,\eta_n\}$ as the filters with the lowest rank in two layers, respectively. Let $a$ and $b$ be positive constants. We assume there is no performance degradation when some filters are pruned if $\sum_{i=1}^{m-x}\xi_i \ge a$ and $\sum_{i=1}^{n-y}\eta_i \ge b$, where $x,y$ are non-negative integers. Suppose $\xi_r$ and $\eta_r$ are selected from $\{\xi_1,...,\xi_m\}$ and $\{\eta_1,...,\eta_n\}$ with equal probability, respectively. Without loss of generality, we suppose $\xi_r=\xi_m$ and $\eta_r=\eta_n$ for simplicity. We compare the performance of five different scenarios: (1) without pruning ($v_o$), (2) pruning randomly-selected filters from the second layer ($v_{sr}$), (3) pruning the lowest-ranked filters in the second layer ($v_{sl}$), (4) pruning the lowest-ranked filters in the first layer ($v_{fl}$), and (5) pruning the globally lowest-ranked filters in both layers ($v_{gl}$).

The $v_o,v_{sr},v_{gl}, v_{fl}, v_{sl}$ can be represented with Eqs. (\ref{eq:vo}) to (\ref{eq:gl}), respectively.
\small
\vskip -0.2in
\begin{equation}
v_o = P(\sum_{i=1}^m\xi_i \ge a) + P(\sum_{i=1}^n\eta_i \ge b)
\label{eq:vo}
\end{equation}
\vskip -0.18in
\begin{equation}
v_{sr} = P(\sum_{i=1}^m\xi_i \ge a) + P(\sum_{i=1}^{n-1}\eta_i \ge b) 
\label{eq:sr}
\end{equation}
\vskip -0.18in
\begin{equation}
v_{sl} = P(\sum_{i=1}^m\xi_i \ge a) + P(\sum_{i=1}^{n}\eta_i - \underline{\eta} \ge b) 
\label{eq:sl}
\end{equation}
\vskip -0.18in
\begin{equation}
v_{fl} = P(\sum_{i=1}^m\xi_i - \underline{\xi} \ge a) + P(\sum_{i=1}^{n}\eta_i \ge b) 
\label{eq:fl}
\end{equation}
\vskip -0.2in
\begin{equation}
\begin{aligned}
v_{gl} &= \frac{m}{m+n}\left[P(\sum_{i=1}^m\xi_i -\underline{\xi} \ge a) + P(\sum_{i=1}^{n}\eta_i \ge b) \right] \\
&+\frac{n}{m+n}\left[P(\sum_{i=1}^m\xi_i \ge a) + P(\sum_{i=1}^{n}\eta_i -\underline{\eta} \ge b) \right]
\end{aligned}
\label{eq:gl}
\end{equation}
\normalsize
Note that $0 \le \eta_n - \underline{\eta} \le \eta_n$, we have

\small
\vskip -0.15in
\begin{equation}
\Rightarrow P(\sum_{i=1}^{n-1}\eta_i \ge b) \le P(\sum_{i=1}^{n}\eta_i - \underline{\eta} \ge b) \le P(\sum_{i=1}^{n}\eta_i \ge b),
\label{eq:relationship}
\end{equation}
\normalsize
which indicates $v_{sr} \le v_{sl} \le v_{o}$.  Moreover, we also have 
\small
\vskip -0.2in
\begin{equation}
P(\sum_{i=1}^{m}\xi_i-\underline{\xi}\ge a) \le P(\sum_{i=1}^{m}\xi_i \ge a),
\label{eq:ximin}
\end{equation}
\vskip -0.15in
\begin{equation}
P(\sum_{i=1}^{n}\eta_i-\underline{\eta}\ge b) \le P(\sum_{i=1}^{n}\eta_i \ge b),
\end{equation}
\normalsize
which indicates $v_{fl} \le v_o$. For the filters in the second layer, we make the following assumptions:
\begin{assumption}
\label{a1}
$\text{$\eta_i$ are uniformly bounded.}~~\exists C_1 > 0,~~~\text{s.t}~~~\mathbb{D}\eta_i \le C_1, i=1,2,\cdots,n$.
\end{assumption}

\begin{assumption}
\label{a2}
$\text{There are at most $C_2\%$ pairs correlated filters.}\\
\#\{(i,j):i\neq j,~i,j=1,2,\cdots,n.~\Cov(\eta_i,\eta_j)\neq 0\} \le C_2n$.
\end{assumption}

\begin{assumption}
\label{a3}
$\text{Filters' contributions are positive.}~~\exists \epsilon_0 > 0,~~~\text{s.t}~~~\mathbb{E}\eta_i \ge \epsilon_0, i=1,2,\cdots,n$.
\end{assumption}
By Chebyshev's inequality, for any real number $\epsilon>0$,
\small
\vskip -0.18in
\begin{equation}
P(\frac{1}{n}|\sum_{i=1}^n(\eta_i-\mathbb{E}\eta_i)|\ge \epsilon) \le \frac{D(\sum_{i=1}^n\eta_i)}{\epsilon^2n^2}.
\label{eq:cheb}
\end{equation}
\normalsize
\vskip -0.15in
From Assumption \ref{a1}, it is obvious that $\Cov(\eta_i,\eta_j)\le\sqrt{D\eta_i \cdot D\eta_j}\le C_1$, together with Assumption \ref{a2},
\small
\vskip -0.18in
\begin{equation*}
\begin{aligned}
D(\sum_{i=1}^n\eta_i) &=\sum_{i=1}^nD\eta_i + \sum_{i \neq j}\Cov(\eta_i,\eta_j) \\
&\le C_1n +C_1C_2n = C_1(1+C_2)n.
\end{aligned}
\end{equation*}
\normalsize
By Eq. \ref{eq:cheb},
\small
\vskip -0.18in
\begin{equation*}
P(\frac{1}{n}|\sum_{i=1}^n(\eta_i-\mathbb{E}\eta_i)|\ge \epsilon) \le \frac{C_1(1+C_2)}{\epsilon^2n} \to 0,
\end{equation*}
\begin{equation}
\frac{1}{n}\sum_{i=1}^n(\eta_i - \mathbb{E}\eta_i) \xrightarrow[]{P} 0.
\label{eq:probapproach0}
\end{equation}
\normalsize
Suppose the number of filters in the second layer $n$ is large enough, say $n>\frac{2b}{\epsilon_0}$, with Assumption \ref{a3}, we have,
\small
\vskip -0.17in
\begin{equation*}
\begin{aligned}
&~~~P(\frac{1}{n}\sum_{i=1}^n(\eta_i-\mathbb{E}\eta_i)>-\frac{\epsilon_0}{2}) = P(\sum_{i=1}^n\eta_i > \sum_{i=1}^n\eta_i - \frac{\epsilon_0}{2}n) \\
&=P(\sum_{i=1}^n\eta_i > \frac{\epsilon_0}{2}n + (\sum_{i=1}^n \mathbb{E}\eta_i - \epsilon_0n)) \\
&\le P(\sum_{i=1}^n\eta_i > \frac{\epsilon_0}{2}n) \le P(\sum_{i=1}^n\eta_i > b).
\end{aligned}
\end{equation*}
\normalsize
Letting $n\to+\infty$ and taking limit, by Eq. \ref{eq:probapproach0} we have
\small
\vskip -0.18in
\begin{equation*}
\lim_{n\to\infty}P(\sum_{i=1}^n\eta_i > b) \ge \lim_{n\to\infty} P(\frac{1}{n}\sum_{i=1}^n(\eta_i-\mathbb{E}\eta_i)>-\frac{\epsilon_0}{2}) = 1.
\end{equation*}
\normalsize
Similarly,
\vskip -0.18in
\small
\begin{equation*}
\lim_{n\to\infty}P(\sum_{i=1}^n\eta_i - \eta_r > b) = \lim_{n\to\infty}P(\sum_{i=1}^n\eta_i - \underline{\eta} > b) = 1,
\end{equation*}
\normalsize
and then we have $v_{sr} \approx v_{sl} \approx v_{o}$. It is also obvious that
\small
\vskip -0.3in
\begin{equation*}
\begin{aligned}
&~~v_{sl} - v_{gl} \\
&\approx \frac{m}{m+n}\left[P(\sum_{i=1}^m\xi_i \ge a) - P(\sum_{i=1}^m\xi_i - \underline{\xi} \ge a)\right] \ge 0,
\end{aligned}
\end{equation*}
\vskip -0.17in
\normalsize

Similarly, we have $v_{gl} - v_{fl} \ge 0$, i.e., $v_{fl}\le v_{gl}\le v_{sl}$.

In summary, we have $v_{fl} \le v_{gl} \le v_{sr} \le v_{sl} \le v_{o}$, which indicates that the network performance after pruning randomly-select filters from large redundant layers is  better than the performance of a network after pruning the least important filters across all layers. 

\section{Empirical Studies and Results}
\label{sec:exp}
In this section, we aim to first verify the Claim as outlined in Section \ref{sec:math} and then demonstrate the superiority of pruning with redundancy reduction with a naive strategy.

\textbf{Setup} We utilize two widely-used architectures (AlexNet \cite{krizhevsky2012ImageNet} and VGG-16 \cite{deng2009ImageNet}) and two benchmark datasets (CIFAR-10 \cite{krizhevsky2009learning}, Birds-200 \cite{welinder2010caltech}) in our experiments for classification purpose. The saliency criterion used in the experiment is Taylor expansion \cite{molchanov2016pruning}, which has achieved the state-of-the-art performance. 
If not explicitly noted, we prune 10 and 50 filters at each run and fine-tune the networks with 100 and 500 updates (with a batch size of 32, and an SGD optimizer with a momentum of 0.9) for AlexNet and VGG-16, respectively. For a fair comparison, we run each experiment five times and report the average results with corresponding standard deviations.

\begin{figure}[ht]
\begin{center}
\centerline{
\subfigure[AlexNet, Layer3]{
\includegraphics[width=0.48\columnwidth]{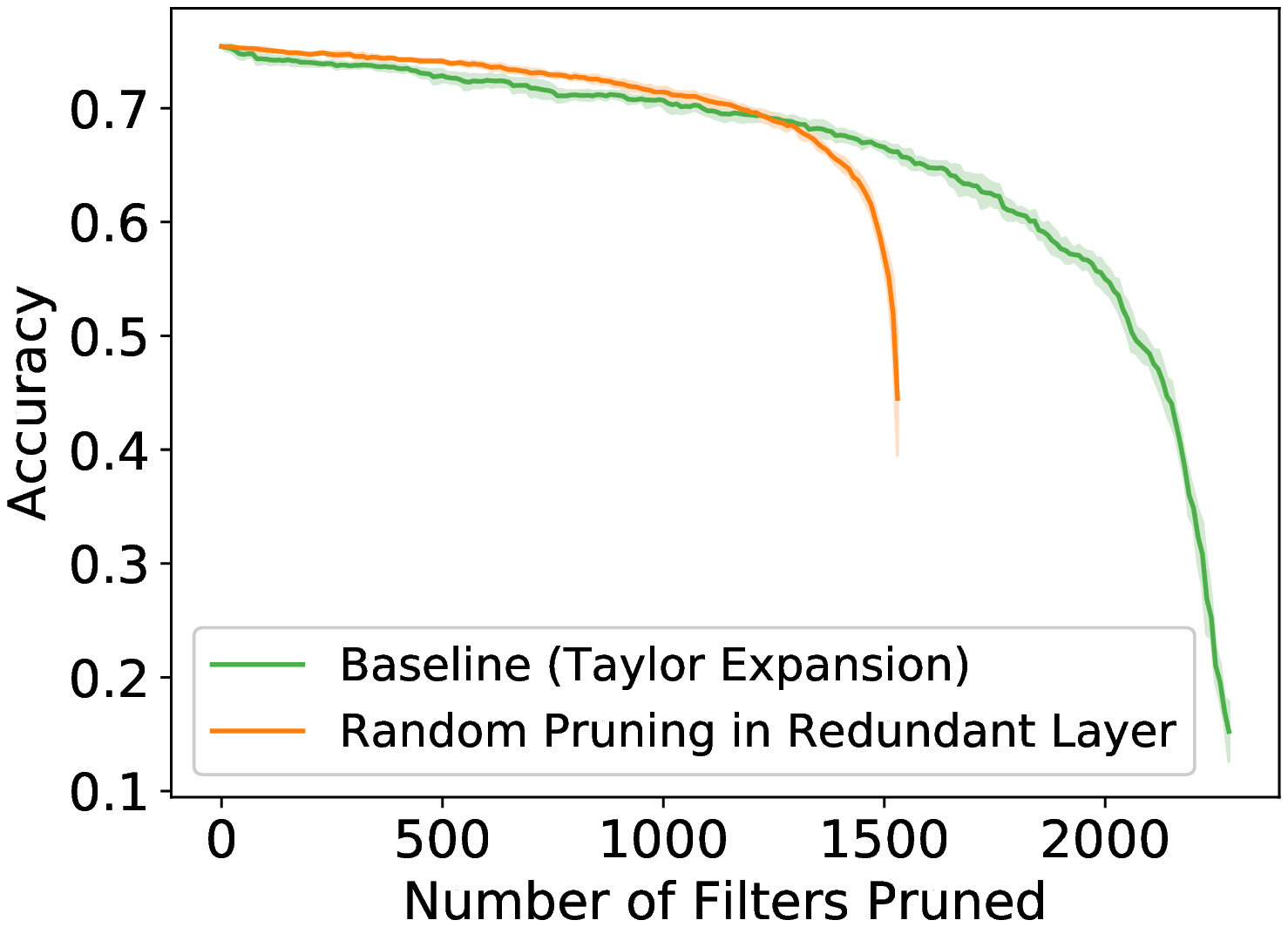}}
\subfigure[AlexNet, Layer4]{
\includegraphics[width=0.48\columnwidth]{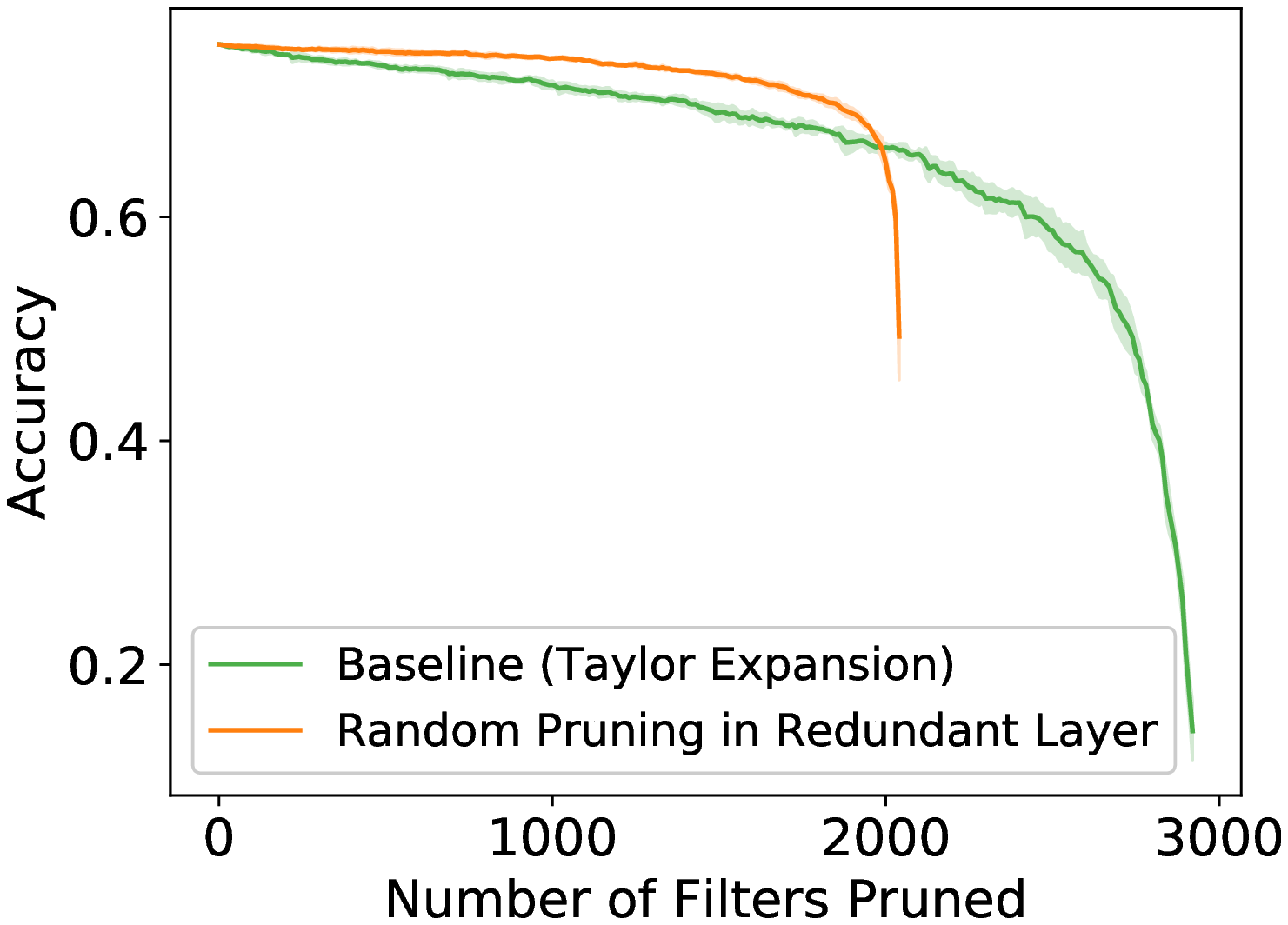}}
}
\vskip -0.15in
\centerline{
\subfigure[VGG-16, Layer9]{
\includegraphics[width=0.48\columnwidth]{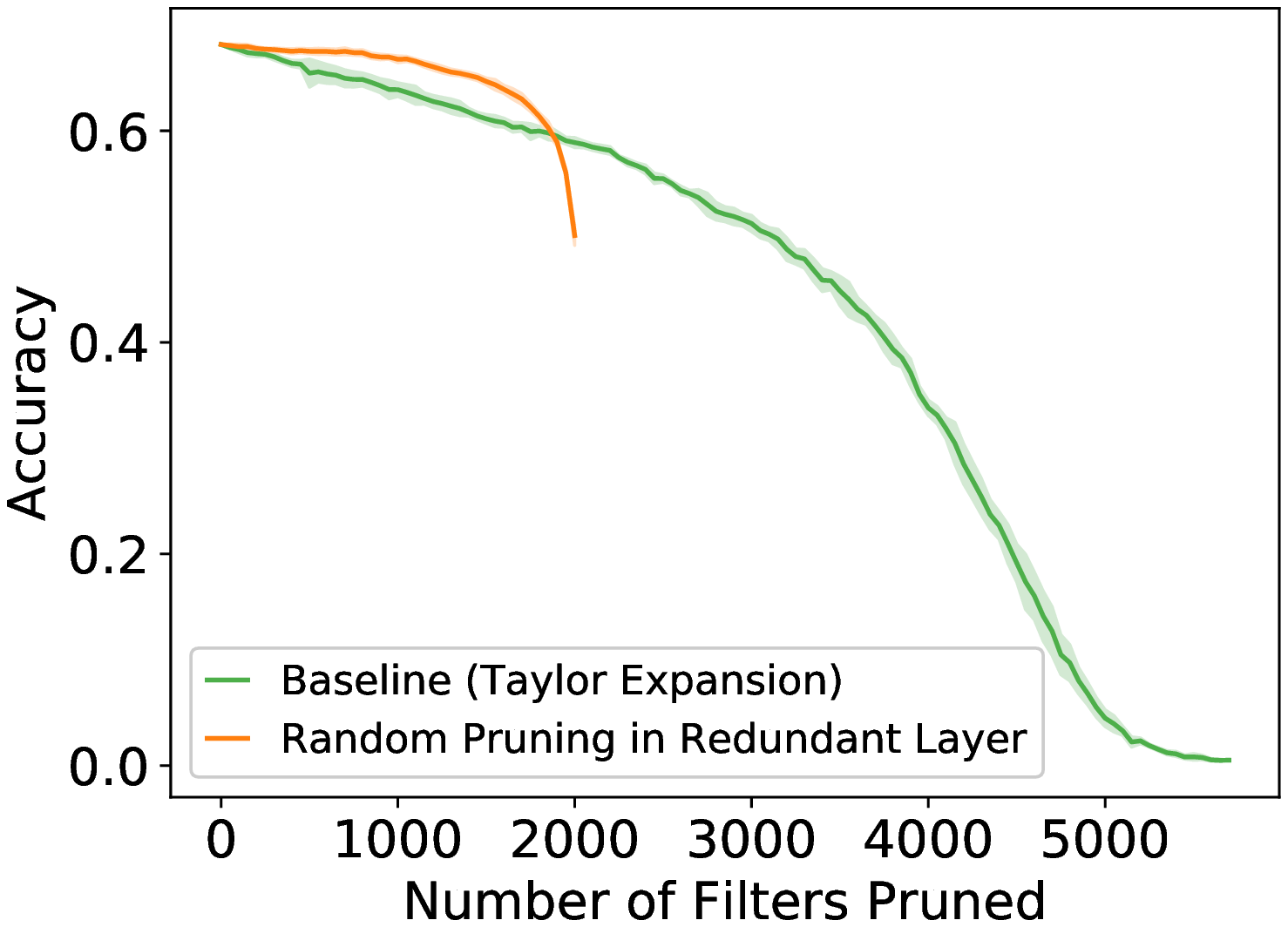}}
\subfigure[VGG-16, Layer12]{
\includegraphics[width=0.48\columnwidth]{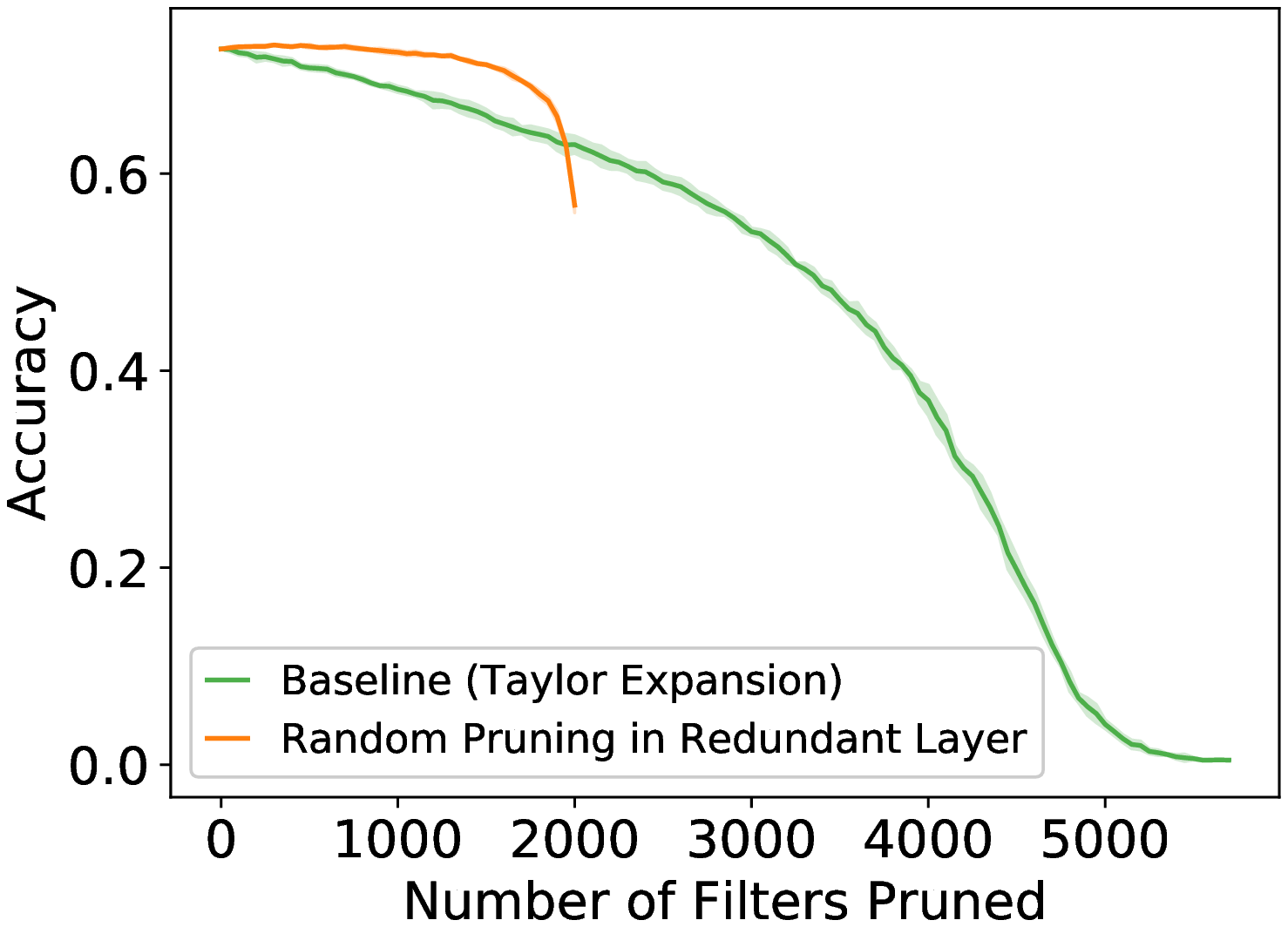}}
}
\vskip -0.15in
\caption{The performance comparison of randomly pruning and Taylor expansion in the case of addition  of redundancy.  Here we show the results with manually-augmented the 3rd and 4th convolutional layers for AlexNet, and 9th and 12th convolutional layers for VGG.}
\label{fig:redundaycy_random_better_than_global_alex3}
\end{center}
\vskip -0.4in
\end{figure}

\textbf{Experiment 1} The derivation in Section \ref{sec:math} considers the situation when large redundancy exists. It mainly focuses on the performance comparison of two pruning strategies: 1): randomly pruning filters from redundant layers, and 2): removing the lowest-ranked filters among all layers. For this purpose, we manually create extremely redundant layers by quadrupling the number of filters in certain layers of benchmark structures. Taylor expansion is used for filter ranking in Strategy 2. The performance comparison of two strategies are shown in Fig.~\ref{fig:redundaycy_random_better_than_global_alex3}. It is clear to see that, consistent with the Claim, Strategy 1 continuously outperforms Strategy 2 until $81\%$, $95\%$, $95\%$ and $95\%$ filters of the predefined redundant layer are pruned in the four cases, respectively. The drastic drop after that is due to the fact that there is a small number of filters left in the predefined redundant layer and redundancy in that layer no longer exists during the pruning process.

\begin{figure}[ht]
\begin{center}
\centerline{
\subfigure[AlexNet, CIFAR-10]{
\includegraphics[width=0.48\columnwidth]{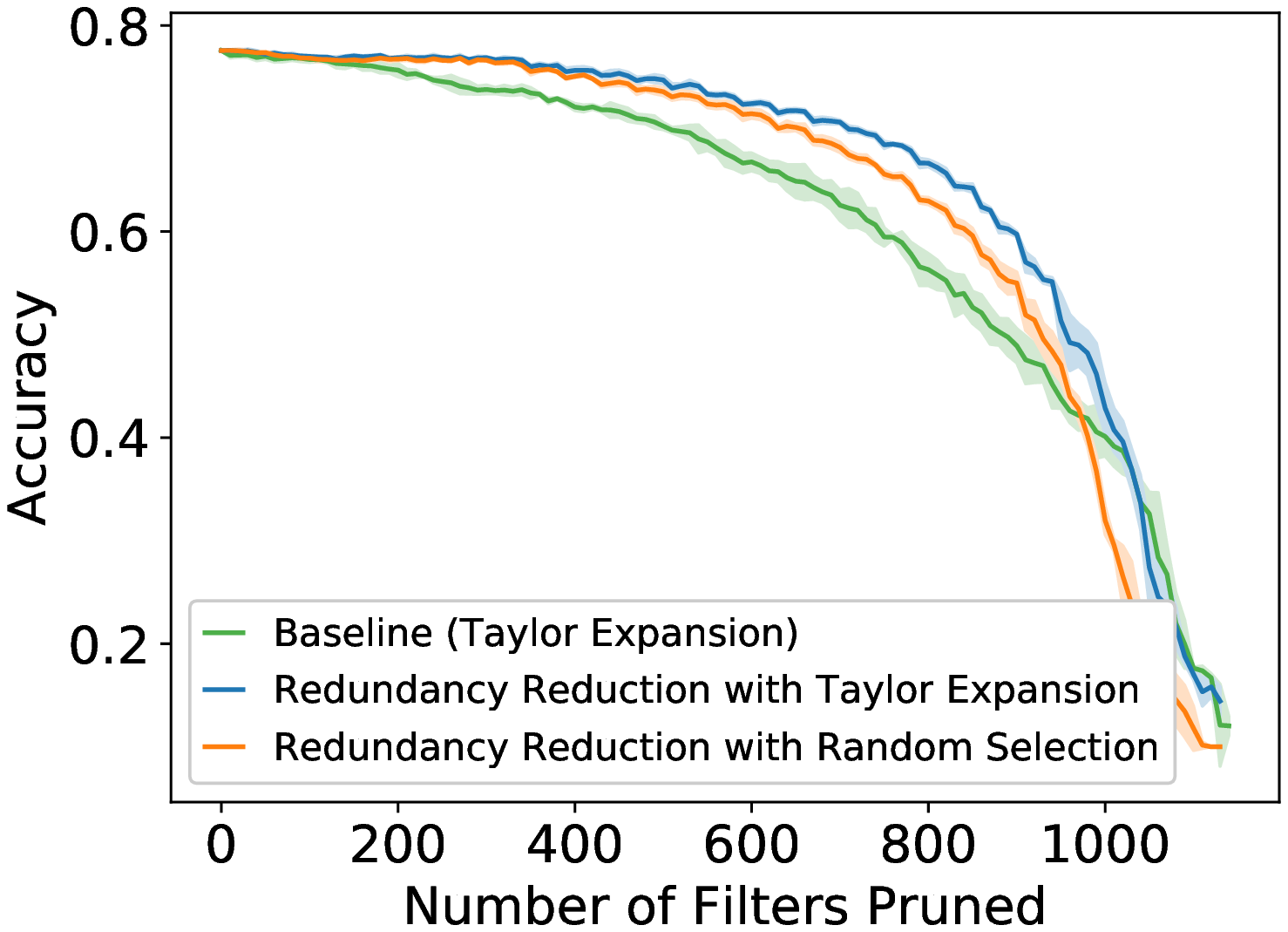}}
\subfigure[VGG-16, Birds-200]{
\includegraphics[width=0.48\columnwidth]{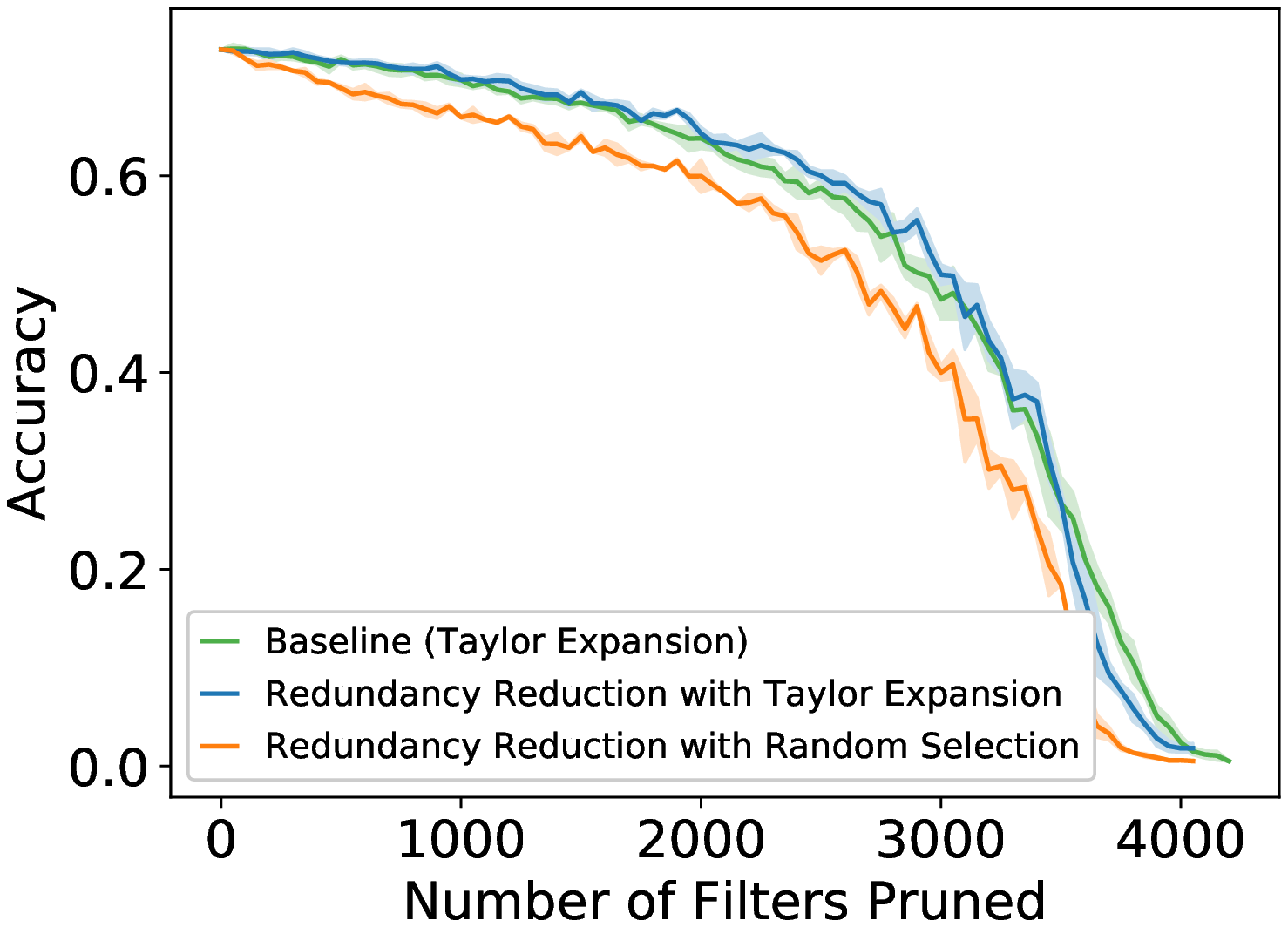}}
}
\vskip -0.15in
\caption{Performance comparison of two redundancy reduction-based strategies and the baseline with Taylor expansion method.}
\label{fig:alex_all_redundancy_strategies}
\end{center}
\vskip -0.3in
\end{figure}

\textbf{Experiment 2} We have confirmed the influence of redundancy in pruning. However, how to find or measure the redundancy for a standard neural network (e.g., original AlexNet or VGG-16 without adding redundancy) remains an open question. Here we use a naive redundancy reduction strategy, i.e., pruning filters from the layers with the most filters at each run to demonstrate the superiority of pruning with redundancy reduction. We consider two redundancy-based strategies here: (1) randomly pruning filters from the layers with the most filters, and (2) pruning the lowest-ranked filters selected by the Taylor expansion approach from the layers with the most filters. The baseline here is still the typical pruning approach with Taylor expansion. The results are illustrated in Fig.~\ref{fig:alex_all_redundancy_strategies}. For AlexNet on CIFAR-10, both of the two redundancy reduction based methods outperform the baseline. For VGG-16 on Birds-200, the redundancy reduction based method with Taylor expansion outperforms the baseline while the redundancy reduction based method with random pruning performs worse than the baseline. The probable reason is that for Birds-200, VGG-16 is a well-distributed structure so that redundancy is evenly distributed in each layer. Overall, such a naive strategy can outperform the state-of-the-art ranking-based pruning approach, indicating that redundancy reduction is a potential direction for future study in this area.
\vskip -0.15in
\section{Conclusion}
\label{sec:conclusion}
In this paper, we formulated the network pruning problem with redundancy reduction. We proved, through both theoretic analysis and empirical studies, random pruning the most redundant layer can outperform importance-based pruning strategies. We also showed that a naive redundancy reduction strategy outperforms well-designed saliency-based pruning approaches, which indicates that exploring the redundancy lying in a neural network is a promising research direction for future network pruning study. 

\bibliography{example_paper}
\bibliographystyle{icml2019}
\end{document}